\documentclass[10pt,twocolumn,letterpaper]{article}

\usepackage{cvpr}              
\usepackage{wrapfig}
\usepackage{bm} 
\usepackage{amsmath}
\usepackage{amsthm}
\usepackage{enumitem}
\theoremstyle{plain}
\newtheorem{theorem}{Theorem}[section]

\theoremstyle{definition}

\theoremstyle{remark}

\definecolor{cvprblue}{rgb}{0.21,0.49,0.74}
\usepackage[pagebackref,breaklinks,colorlinks,allcolors=cvprblue]{hyperref}
\usepackage{comment}

\def\paperID{10911} 
\def\confName{CVPR}
\def\confYear{2025}

\title{Detecting Out-of-distribution through the Lens of Neural Collapse}

\author{Litian Liu\\
MIT\\
{\tt\small litianl@mit.edu}
\and
Yao Qin\\
UC Santa Barbara\\
{\tt\small yaoqin@ucsb.edu}
}

\begin{document}
\maketitle
\begin{abstract}
Out-of-Distribution (OOD) detection is critical for safe deployment; however, existing detectors often struggle to generalize across datasets of varying scales and model architectures, and some can incur high computational costs in real-world applications. Inspired by the phenomenon of Neural Collapse, we propose a versatile and efficient OOD detection method. Specifically, we re-characterize prior observations that in-distribution (ID) samples form clusters, demonstrating that, with appropriate centering, these clusters align closely with model weight vectors. Additionally, we reveal that ID features tend to expand into a simplex Equiangular Tight Frame, explaining the common observation that ID features are situated farther from the origin than OOD features. Incorporating both insights from Neural Collapse, our OOD detector leverages feature proximity to weight vectors and complements this approach by using feature norms to effectively filter out OOD samples. Extensive experiments on off-the-shelf models demonstrate the robustness of our OOD detector across diverse scenarios, mitigating generalization discrepancies and enhancing overall performance, with inference latency comparable to that of the basic softmax-confidence detector.
Code is available at:
\url{https://github.com/litianliu/NCI-OOD}.
\end{abstract}    
\vspace{-5mm}
\section{Introduction}
\label{sec:intro}

Machine learning models deployed in practice will inevitably encounter samples that deviate from the training distribution.
As a classifier cannot make meaningful predictions on test samples that belong to classes unseen during training, it is important to actively detect and handle Out-of-Distribution (OOD) samples.
Considering the diverse and oftentimes time-critical application scenarios, an OOD detector should be computationally efficient and can effectively generalize across various scenarios.

In this work, we focus on \emph{post-hoc} methods, which address OOD detection independently of the training process. 
One line of prior work designs OOD scores over model output space \citep{djurisic2022extremely, hendrycks2019scaling,liang2018enhancing, liu2020energy,  sun2021react, sun2022dice} 
and another line of work focuses on the feature space, where OOD samples are observed to deviate from clusters of ID samples \citep{lee2018simple, mahalanobis2018generalized, sun2022out, tack2020csi}. 
While existing research has made strides in OOD detection, they still face two major challenges: 1) maintaining detection effectiveness across different scenarios, and 2) ensuring computational efficiency for real-world deployment. 
For example, both output space and feature space methods suffer from performance discrepancy across different classification tasks, as shown in Table~\ref{tab:main_full}~(a). 
Specifically, strong algorithms on detecting OOD examples for CIFAR-10 \citep{krizhevsky2009learning} perform suboptimally in detecting OOD examples for ImageNet \citep{deng2009imagenet}, and vice versa. 
No existing algorithm can simultaneously rank in the top three across two benchmarks, leading to sub-optimal average performance as shown in Table~\ref{tab:main_full}~(b).
Furthermore, feature-space methods that rely on similarity measures to characterize ID clustering raise efficiency concerns. For example, \cite{sun2022out} records training features and evaluates OOD-ness based on the k-th nearest neighbor distance to these features, while \cite{liu2024fast} demonstrates that dependency on such auxiliary models increases computational costs, posing challenges for time-sensitive applications. This highlights the need to move \emph{beyond} feature-wise similarity and establish a more structured reference to characterize ID clustering efficiently.

\begin{figure}{}
\vspace{0mm}
\centering
\includegraphics[width=0.43\textwidth]{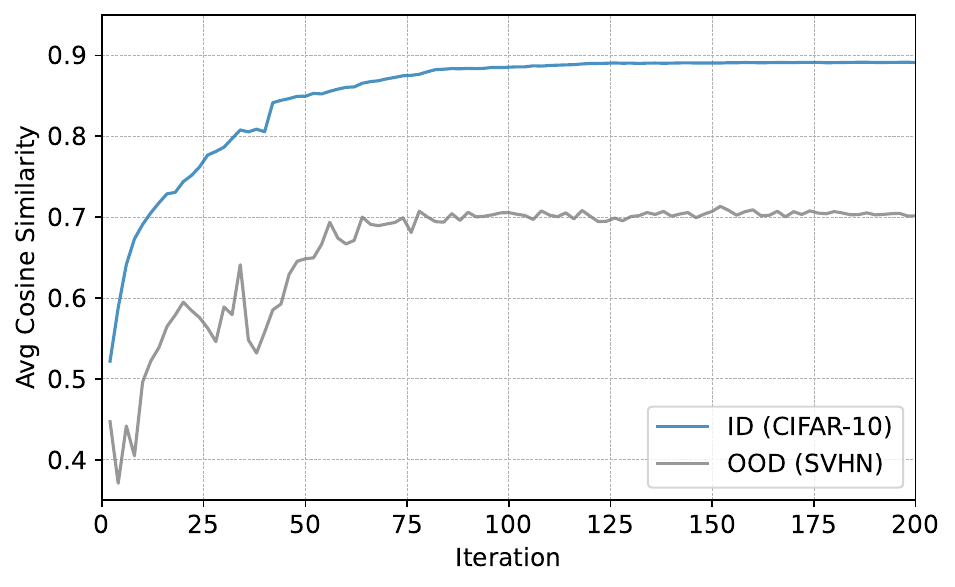}
\caption{\textbf{Centered ID samples tend to cluster near the predicted class weight vectors}, which are the last-layer weights of the corresponding class, as indicated by higher average cosine similarity (Equation~\ref{eq:avgCos}) than OOD.
This observation, inspired by the trend of Neural Collapse, emerges early in the training of this CIFAR-10 ResNet-18 classifier, with OOD set SVHN.
}\label{fig:evolution}
\vspace{-2mm}
\end{figure}

\begin{figure*}[t]
\vspace{-5mm}
\begin{center}
\centerline{\includegraphics[width=0.98\textwidth]{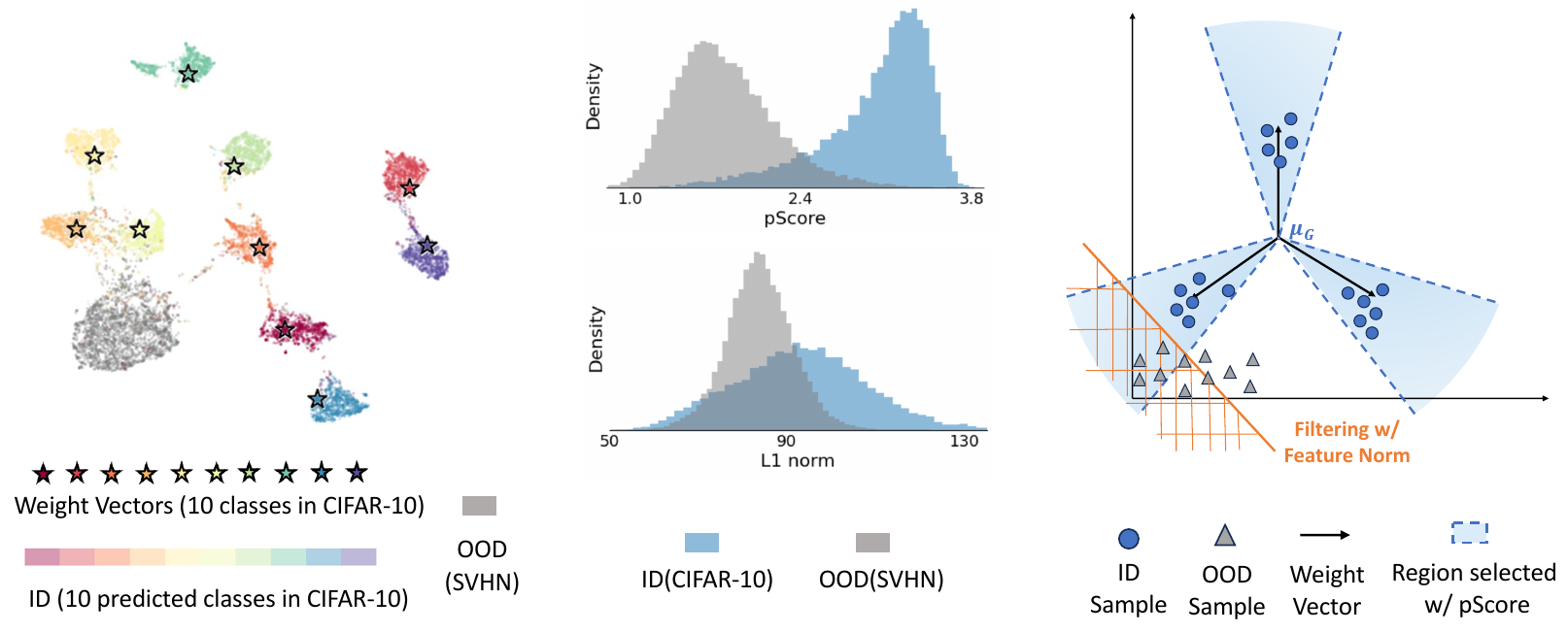}}
\end{center} 
\vspace{-8.5mm}
\caption{
\textbf{Framework Illustration.} 
\textit{Left:}
On the penultimate layer, the centered 
ID clusters reside near their predicted class weight vectors (marked by stars) while OOD samples reside separated, as shown by UMAP. 
\textit{Middle:}
ID and OOD samples are separated by $\mathtt{pScore}$ (Equation~\ref{eq:prox}), which measures feature proximity to weight vectors. 
Also, ID samples tend to be further from the origin, illustrated with $\mathtt{L1}$ norms. 
\textit{Right:}
ID samples cluster near a simplex Equiangular Tight Framework, illustrated with black arrows denoting weight vectors.   
We detect OOD by thresholding on $\mathtt{pScore}$, selecting blue-shaded hypercones centered at weight vectors, with OOD samples outside these areas.
We also filter OOD samples characterized by smaller feature norms.  
\textit{Left} \& \textit{Middle} present a practical \emph{off-the-shelf} CIFAR-10 ResNet-18 classifier with OOD set SVHN. 
\textit{Right} depicts our scheme on a three-class classifier with 2D penultimate space. 
}\label{fig:demo-score}
\end{figure*}

To this end, we revisit the well-established observation that penultimate-layer ID features tend to form clusters, while OOD features reside apart \cite{lee2018simple, sun2022out, tack2020csi}, and take it a step further by asking:
\begin{center}
\emph{Where do features of ID samples form clusters?}
\end{center}

Leveraging \textit{Neural Collapse} \citep{papyan2020prevalence}, we first examine the theoretical landscape with models trained beyond zero training error. 
We first show that, as a deterministic effect of \textit{Neural Collapse}\citep{papyan2020prevalence} penultimate-layer features of training samples converge toward the \emph{weight vectors of the predicted class}—the last-layer weights of the corresponding class— after being {centered} by training feature mean.
Additionally, Neural Collapse demonstrates that training features conform to a simplex Equiangular Tight Frame (ETF) (Equation~\ref{eq:etf}), representing the maximum achievable separation between equiangular vectors. This spatial structure, illustrated in Figure~\ref{fig:demo-score} \textit{Right}, requires features to be sufficiently distant from the origin.

Building on the insights from the theoretical convergence landscape, we propose two hypotheses about the geometric structure of ID clusters and OOD samples in practical models. 
Specifically, for ID test samples drawn from the same distribution as training samples, we expect a similar {trend} of clustering behavior toward the weight vectors and the ETF structure. In contrast, OOD samples, which are not part of the training process, lack the alignment with weight vectors and the spatial expansion necessary to form an ETF. Consequently, the model is unlikely to align unseen OOD features with weight vectors or place them far enough from the origin to exhibit ETF-like behavior. Although complete Neural Collapse requires strict conditions, such as prolonged training, its convergence trend and corresponding geometric patterns of ID features are consistently observed during training across diverse architectures and classification tasks (see Appendix~\ref{app:nc_prevelance} and \citep{he2023law}). 
Thus, our hypotheses regarding ID clusters and OOD samples are expected to hold without requiring complete Neural Collapse.

To validate the hypotheses, we trace a model's training stages in Figure~\ref{fig:evolution}, where centered ID samples consistently cluster closer to the weight vectors than OOD samples.
By epoch 50, such a pattern already enables effective OOD detection for SVHN samples
\footnote{Our method (Section~\ref{sec:algorithm}) achieves an AUROC of 94.44, outperforming the 91.27 softmax-confidence baseline (AUROC defined in Section~\ref{sec:experiments}).}. 
This shows that the trend of Neural Collapse emerges early, validating the practical effectiveness of our hypothesis before full collapse in mature models.
We reinforce this observation with a UMAP \citep{mcinnes2018umap} visualization of an \emph{off-the-shelf} CIFAR-10 classifier with a ResNet-18 backbone, shown in Figure~\ref{fig:demo-score}~\emph{Left}. 
ID features cluster near the predicted class weights (marked by stars), while OOD features remain distant.
Additionally, as \citep{zhu2021geometric} shows that weight vectors form an ETF, our observations also support our second hypothesis: ID features are driven to structure the ETF, while OOD features lack the incentive to expand in space.
This explains the well-established observation \citep{tack2020csi, huang2021importance, sun2022out} that OOD features tend to reside closer to the origin, providing an alternative to the model confidence interpretation in \cite{park2023understanding}. 

Based on our understanding, we design an efficient and versatile OOD detector. 
We first leverage feature proximity to the weight vectors to characterize ID clustering, effectively incorporating class-specific information and reducing the computational cost. 
Specifically, we define an angle-based proximity score as the norm of the projection of the weight vector of the predicted class onto the centered sample feature.
As shown in Figure~\ref{fig:demo-score} \textit{Middle}, our proximity score can effectively separate ID/OOD.
A higher score indicates closer proximity and a lower chance of OOD-ness.
Geometrically, thresholding on the score selects hyper-cones centered at the weight vector, 
as illustrated in Figure~\ref{fig:demo-score} \textit{Right}. 
Complementing the proximity score's contingency on ID clustering, we also consider feature distance to the origin. 
Specifically, ID features tend to reside further from the origin as they expand in space to form an ETF, whereas OOD features tend to reside near the origin, as illustrated by Figure~\ref{fig:demo-score} \textit{Right}.
Using the L1 norm as an example metric 
, we observe that ID features can be separated from OOD features, as supported by Figure~\ref{fig:demo-score} \textit{Middle}.
Combining both aspects, we propose \textbf{N}eural \textbf{C}ollapse \textbf{I}nspired OOD Detector ($\mathtt{NCI}$). 

Notably, prior methods, e.g., KNN~\cite{sun2022out}, focus on ID clustering, without explicitly incorporating feature distance to the origin. 
Such approaches fall short in scenarios like ImageNet benchmarks but yield superior performance in CIFAR-10 benchmarks in Table~\ref{tab:main}.
Conversely, methods such as Energy~\cite{liu2020energy}, Energy-based ASH~\cite{djurisic2022extremely}, and, Energy-based Scale~\cite{xu2023scaling} inherently utilize feature distance to the origin by considering log-sum-exp of logits, yet largely overlook ID clustering. 
These approaches excel in scenarios like ImageNet, but perform sub-optimally in others, e.g., CIFAR-10.
Through the lens of Neural Collapse, we explain, connect, and complete a wide range of prior methods under a holistic view, resulting in reduced latency and generalization discrepancies.  

We summarize our main contributions below:

\begin{itemize}
    \item 
    \textbf{Understanding and Observation:} 
    By analyzing ID clustering through the trend of Neural Collapse, we discover that penultimate-layer features of ID test examples cluster around the weight vectors of the predicted class after being centered by the training feature mean. Additionally, we explain that ID features tend to lie farther from the origin due to Neural Collapse, where they form a simplex Equiangular Tight Frame (ETF) structure. Notably, our findings hold without requiring full Neural Collapse convergence.
    \item \textbf{OOD Detector:}
    We leverage feature proximity to the weight vectors of predicted classes for OOD detection, integrating class-specific information. 
    Complementary to feature clustering, we propose to detect OOD samples by thresholding the feature distance to the origin.
    \item \textbf{Experimental Analysis:}
    We evaluate $\mathtt{NCI}$ across diverse classification tasks (CIFAR-10, CIFAR-100, ImageNet) and model architectures (ResNet, DenseNet \cite{huang2017densely}, ViT \cite{dosovitskiy2020image}, Swin \cite{liu2022swin}).
    Rather than focusing on \emph{individual} benchmarks, $\mathtt{NCI}$ reduces the generalization discrepancies and improves the \emph{overall} effectiveness.
    We also show that incorporating weight vectors not only significantly improves upon prior clustering-based methods but also reduces latency to a minimal level, matching the basic \emph{softmax-confidence} detector.
\end{itemize}

\section{Problem Statement}
\label{sec:problem}

We consider a data space $\mathcal{X}$, a class set $\mathcal{C}$, and a classifier $f:\mathcal{X} \rightarrow \mathcal{C}$, which is trained on samples \emph{i.i.d.} drawn from joint distribution $\mathbb{P}_{\mathcal{X}\mathcal{C}}$.
We denote the marginal distribution of $\mathbb{P}_{\mathcal{X}\mathcal{C}}$ on $\mathcal{X}$ as $\mathbb{P}^{in}$. 
And samples drawn from $\mathbb{P}^{in}$ are In-Distribution (ID) samples. 
In practice, the classifier $f$ may encounter $\bm{x} \in \mathcal{X}$ yet is not drawn from $\mathbb{P}^{in}$.
We say such samples are Out-of-Distribution (OOD). 

In this work, we focus on detecting OOD samples from \emph{classes unseen during training}, for which the classifiers cannot make meaningful predictions.
The OOD detector $D: \mathcal{X} \rightarrow \{\text{ID}, \text{OOD}\}$ is commonly constructed as: $D(\bm{x}) = \begin{cases} \text{ID} & \text{if } s(\bm{x}) \geq \tau \\ \text{OOD} & \text{if } s(\bm{x}) < \tau \end{cases}$, where $s: \mathcal{X} \rightarrow \mathbb{R}$ is a score function of design and $\tau$ is the threshold.
Considering the diverse application scenarios, an ideal OOD detector should be efficient and generalizable.
Thus, we leverage insights from Neural Collapse to achieve reduced computational costs and minimize generalization discrepancies. 
\vspace{-1mm}
\section{OOD Detection through the Lens of Neural Collapse}\label{sec:methodology}
\vspace{-1mm}

In this section, we re-examine the observation in \cite{lee2018simple, sun2022out} that ID features tend to form clusters while OOD features deviate from the clusters.
We suggest that the clustering phenomenon can reflect the trend of the Neural Collapse \citep{papyan2020prevalence} in practical models.
From our understanding, we develop a \emph{post-hoc} OOD detector with enhanced efficiency and effectiveness. 

\subsection{Neural Collapse: Convergence Landscape}\label{sec:nc_theory}

Neural Collapse, first observed in \cite{papyan2020prevalence}, occurs on the penultimate layer across canonical classification settings. 
To formally introduce the concept, we use $\bm{h}_c^i$ to denote the penultimate layer feature of the $i_{th}$ training sample with ground truth / predicted label $c$, Neural Collapse is framed in relation to 
\begin{itemize}
    \item feature global mean, $\bm{\mu}_G = \mathrm{Ave}_{i,c} \left(\bm{h}_{c}^i\right)$, where $\mathrm{Ave}\left(\cdot\right)$ is the average operation;
    \item feature class means, $\bm{\mu}_c = \mathrm{Ave}_{i} \left(\bm{h}_{c}^i\right), \ \forall c \in \mathcal{C}$;
    \item within-class covariance, 
    $$\bm{\Sigma}_W = \mathrm{Ave}_{i,c}\left((\bm{h}_{c}^i - \bm{\mu}_c)(\bm{h}_{c}^i - \bm{\mu}_c)^T\right);$$
    \item between-class covariance,
    $$\bm{\Sigma}_B = \mathrm{Ave}_{c}\left((\bm{\mu}_c - \bm{\mu}_G)(\bm{\mu}_{c} - \bm{\mu}_G)^T)\right);$$
    \item linear classification head, i.e. the last layer of the NN, $\arg \max_{c \in \mathcal{C}} \bm{w}_c^T\bm{h} + b_c$, where $\bm{w}_c$ and $b_c$ are parameters corresponding to class $c$. 
\end{itemize}
Neural Collapse comprises four inter-related behaviors: 

\textbf{(NC1) Within-class variability collapse: } $\bm{\Sigma}_W \rightarrow \bm{0}$

\textbf{(NC2) Convergence to a simplex Equiangular Tight Frame (ETF): }
\begin{equation}\label{eq:etf}
\begin{aligned}
& |\|\bm{\mu}_c - \bm{\mu}_G \|_2 - \|\bm{\mu}_{c'} - \bm{\mu}_G \|_2 | \rightarrow 0, \ \forall \ c, \ c' \\
& \frac{(\bm{\mu}_c - \bm{\mu}_G)^T(\bm{\mu}_{c'} - \bm{\mu}_G)}{\|\bm{\mu}_c - \bm{\mu}_G \|_2 \|\bm{\mu}_{c'} - \bm{\mu}_G \|_2} \rightarrow \frac{|\mathcal{C}|}{|\mathcal{C}|-1}\delta_{c, c'} - \frac{1}{|\mathcal{C}|-1}  
\end{aligned}
\end{equation}
where $\delta_{c, c'}$ is the Kronecker delta symbol.

\textbf{(NC3) Convergence to self-duality: }
\vspace{-1mm}
\[
\frac{\bm{w}_c}{\|\bm{w}_c\|_2} - \frac{\bm{\mu}_c - \bm{\mu}_G}{\|\bm{\mu}_c - \bm{\mu}_G \|_2} \rightarrow \bm{0} 
\]

\vspace{-1mm}
\textbf{(NC4) Simplification to nearest class center: }
\begin{align*}
\arg \max_{c \in \mathcal{C}} \bm{w}_c^T\bm{h} + b_c  \rightarrow arg \min_{c \in \mathcal{C}} \|\bm{h} - \bm{\mu}_c\|_2
\end{align*}

\vspace{-2mm}
We first remark on \textbf{(NC2)} that an ETF achieves the maximum separation possible for globally centered equiangular vectors \cite{papyan2020prevalence} and extends in space, as visualized in Figure~\ref{fig:demo-score} \textit{Right}. 
Since training features converge towards an ETF, they need to have sufficient norms to accommodate the spatial arrangement. 

We next build on \textbf{(NC1)} and \textbf{(NC3)} to demonstrate that training features converge towards the weight vectors of the linear classification head, up to a scaling factor. 

\begin{theorem}\label{thm:1}
\textbf{(NC1)} and \textbf{(NC3)} imply that for any sample $i$ and its predicted class $c$, we have 
\vspace{-1mm}
\begin{equation}\label{eq:thm1}
    (\bm{h}_{c}^i - \bm{\mu}_G) \rightarrow  \lambda \bm{w}_c,  
\vspace{-0.5mm}
\end{equation}
\vspace{-0.5mm}
where \( \displaystyle \lambda = \frac{\| \bm{\mu}_{c} - \bm{\mu}_G\|_2}{\|\bm{w}_c\|_2} \) in the Terminal Phase of Training. 


\begin{proof}
Considering that $(\bm{h}_{c}^i - \bm{\mu}_c)(\bm{h}_{c}^i - \bm{\mu}_c)^T$ is positive semi-definite for any $i$ and $c$.
$\bm{\Sigma}_W \rightarrow \bm{0}$ thus implies $(\bm{h}_{c}^i - \bm{\mu}_c)(\bm{h}_{c}^i - \bm{\mu}_c)^T \rightarrow \bm{0}$ and $\bm{h}_{c}^i - \bm{\mu}_c \rightarrow \bm{0}, \ \forall i, c$. 
With algebraic manipulations, we have
\vspace{-1.5mm}
\begin{equation}\label{pf: step_2}
\frac{\bm{h}_{c}^i - \bm{\mu_G}}{\| \bm{\mu}_c - \bm{\mu_G}\|_2} - \frac{\bm{\mu}_c - \bm{\mu_G}}{\| \bm{\mu}_c - \bm{\mu_G}\|_2} \rightarrow \bm{0}, \ \forall i, c
\end{equation}
Applying the triangle inequality, we have
\begin{equation}\label{eq:triangle}
\begin{aligned}
\scriptstyle
& |\frac{\bm{h}_{c}^i - \bm{\mu_G}}{\| \bm{\mu}_c - \bm{\mu_G}\|_2} - \frac{\bm{w}_c}{\|\bm{w}_c\|_2} | \leq \\
& |\frac{\bm{h}_{c}^i - \bm{\mu_G}}{\| \bm{\mu}_{c} - \bm{\mu_G}\|_2} - \frac{\bm{\mu}_c - \bm{\mu_G}}{\| \bm{\mu}_c - \bm{\mu_G}\|_2}| + |\frac{\bm{w}_c}{\|\bm{w}_c\|_2} - \frac{\bm{\mu}_c - \bm{\mu}_G}{\|\bm{\mu}_c - \bm{\mu}_G \|_2}|.
\end{aligned}
\end{equation}
Since both terms on the RHS converge to $\bm{0}$, as demonstrated by \eqref{pf: step_2} and \textbf{(NC3)}, it follows that the LHS also converges to $\bm{0}$.
\vspace{-1mm}
\end{proof}
\end{theorem}

\subsection{Trend of Neural Collapse \& Geometric Structure of the ID Clusters}

While the complete collapse occurs during the Terminal Phase of Training (TPT) where training error vanishes and the training loss is trained towards zero, it is observed in \cite{he2023law} that the trend of Neural Collapse establishes in the early stages of training.
We thus suggest that the clustering behavior of ID features observed in \emph{off-the-shelf} models reflects a trend of Neural Collapse, corresponding to the within-class variability collapse $\textbf{(NC1)}$.  
In light of this, we use the landscape of Neural Collapse revealed in Theorem~\ref{thm:1} and \textbf{(NC2)} to examine the geometry of ID clusters. 

Following our discussion in Section~\ref{sec:intro}, we we hypothesize and validate with pre-trained models that 
(1) ID features tend to cluster closer to the weight vectors compared to OOD features;
(2) ID clusters tend to reside further from the origin, as necessitated by their spatial structure. 
Particularly, Neural Collapse highlights the importance of \emph{centering} for charaterizing ID clustering with weight vectors.  
We thus validate our hypothesis in a CIFAR-10 classifier with ResNet-18 backbone in Figure~\ref{fig:evolution}, computing over the ID set (CIFAR-10) and OOD set (SVHN) the average cosine similarity between the centered feature $\bm{h}^i - \bm{\mu}_G$ and the weight vector $\bm{w}_c$ of the predicted class $c$, i.e.,
\begin{equation}\label{eq:avgCos}
    \mathrm{Ave}_i \left( \frac{(\bm{h}^i - \bm{\mu}_G) \cdot \bm{w}_c}{\| \bm{h}^i - \bm{\mu}_G\|_2 \| \bm{w}_c \|_2} \right)
\end{equation}

\subsection{Out-of-Distribution Detection}\label{sec:algorithm}

Based on our understanding, we design an efficient and versatile OOD detector.
Specifically, we propose to detect OOD based on feature proximity to the weight vectors of the predicted class.
For the proximity metric, we avoid Euclidean-based metrics as they require estimating the scaling factor $\lambda$ in Equation~\ref{eq:thm1}. 
This estimation tends to be imprecise for general classifiers which may cease training prior to convergence, resulting in suboptimal performance of Euclidean-based metrics shown in Appendix~\ref{app:alter_metric}. 
Instead, we design an angle-based metric, adjusted for class-wise difference. 
Specifically, we propose to quantify the proximity as the norm of projection of the weight vector $\bm{w}_c$ onto the centered feature $\bm{h} - \bm{\mu}_G$, where $c$ corresponds to the predicted class, i.e., 
\begin{equation}\label{eq:prox}
    \mathtt{pScore} = cos(\bm{w}_c, \bm{h} - \bm{\mu}_G)\|\bm{w}_c\|_2, 
\end{equation}
where $cos(\bm{w}_c, \bm{h} - \bm{\mu}_G) = \frac{(\bm{h} - \bm{\mu}_G) \cdot \bm{w}_c}{\| \bm{h} - \bm{\mu}_G\|_2\|\bm{w}_c\|_2}$.
A higher $\mathtt{pScore}$ indicates closer proximity to the weight vector and thus a lower chance of OOD-ness. 
Geometrically, thresholding on $\mathtt{pScore}$ selects infinite hyper-cones centered at the weight vectors, as illustrated in Figure~\ref{fig:demo-score}~\textit{Right}.
Within the same predicted class, $\mathtt{pScore}$ is proportional to the cosine similarity.
Across different classes, $\mathtt{pScore}$ adapts to class-wise difference by selecting wider hyper-cones for classes with larger weight vectors, which tend to have larger decision regions.
As shown in Appendix~\ref{app:alter_metric}, our $\mathtt{pScore}$ with class-wise adjustment outperforms vanilla cosine similarity. 
Notably, our $\mathtt{pScore}$ incorporates class-specific information into characterizing ID clustering by using the weight vectors of the predicted class. 
This brings in additional gain in detection effectiveness, as we shall see in Section~\ref{sec:experiments}. 

While $\mathtt{pScore}$ enhances efficiency and effectiveness, its performance is intrinsically contingent on the strength of ID clustering. 
Such contingency, widely exhibited by clustering-based methods \cite{lee2018simple, sun2022out, tack2020csi}, poses challenges on classifiers with less pronounced ID clustering, such as ImageNet ResNet-50 in Section~\ref{sec:versatile}.
To mitigate such discrepancy,  
we complement $\mathtt{pScore}$ by using the distance of ID clusters to the origin. 
Specifically, we enhance our proximity score by incorporating feature norms to filter out OOD near the origin, as illustrated in Figure~\ref{fig:demo-score}~\emph{Right}. 
Taking $\mathtt{L1}$ norm as an example, we define our detection score as $\mathtt{pScore} + \alpha \|\bm{h}\|_1$, where $\alpha$ controls the filtering strength and can be selected from a validation set as detailed in Section~\ref{sec:experiments}. 
We discuss the effect of different orders of $p$-norm in Section~\ref{sec:ablation}. 
Thresholding on the detection score, we have \textbf{N}eural \textbf{C}ollapse \textbf{I}nspired OOD Detector ($\mathtt{NCI}$):
A lower score indicates a higher chance of OOD-ness.

$\mathtt{NCI}$ 
has \(O(P)\) complexity, where \(P\) is the penultimate layer dimension. 
The complexity theoretically ensures computational scalability of $\mathtt{NCI}$ on large models. 
Empirically, $\mathtt{NCI}$ maintains inference latency comparable to the vanilla \emph{softmax-confidence} detector, as shown in Section~\ref{sec:experiments}. 
\vspace{-1mm}

\section{Experiments}\label{sec:experiments}
\begin{table*}[t]
    \vspace{-3mm}
    \caption{$\mathtt{NCI}$ reduces discrepencies and improves \textbf{overall performance} on CIFAR-10 and ImageNet benchmarks with \textbf{minimal latency}.
    CIFAR-10 uses ResNet-18 and ImageNet uses ResNet-50. 
    }\label{tab:main_full}
    \vspace{-3mm}
    \centering
    \begin{subtable}{\linewidth}
        \begin{center}        \includegraphics[width=0.99\textwidth]{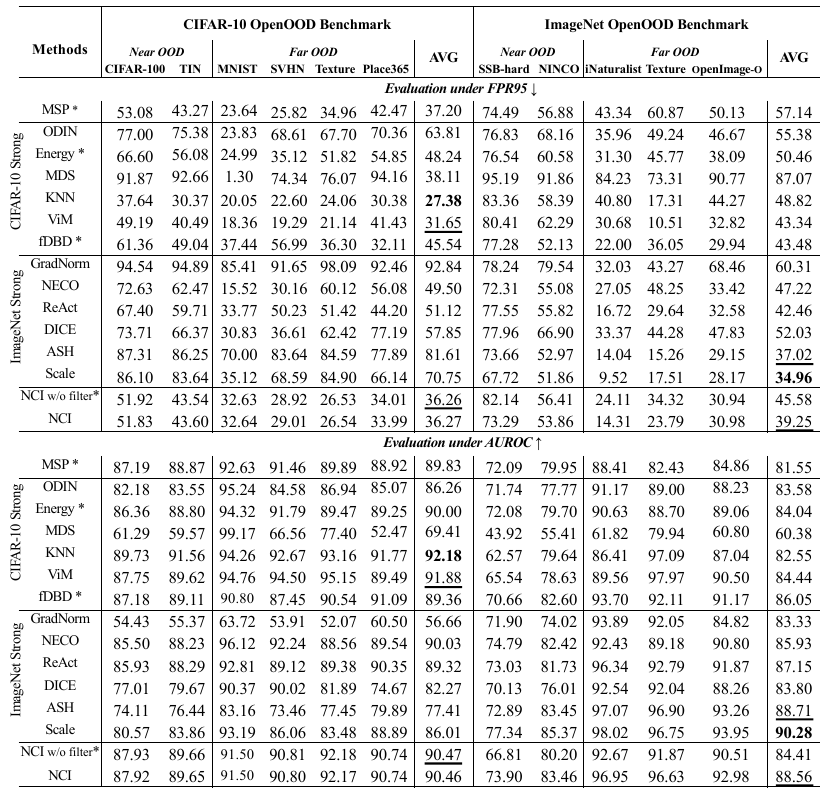} 
    \end{center}
        \centering
        \vspace{-3mm}
        \begin{minipage}{\linewidth}
        \centering
        \caption{
        $\mathtt{NCI}$ ranks \textbf{top-three} in both benchmarks, while baselines show greater variability.    
        $\uparrow$ and $\downarrow$ denotes better performance.
        \textbf{Bold} marks best, \underline{underline} 2nd / 3rd.
        Methods with * are hyperparameter-free.
        Scores, except for the most recent baselines -- $\mathtt{fDBD}$, $\mathtt{NECO}$, $\mathtt{ASH}$, $\mathtt{Scale}$ -- are from OpenOOD \cite{zhang2023openood}.
        }\label{tab:main}
        \end{minipage}
    \end{subtable}
    \begin{subtable}{\linewidth}
        \begin{center}        \includegraphics[width=0.82\textwidth]{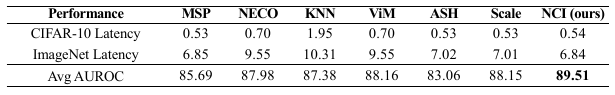} 
\end{center}
        \vspace{-1mm}
        \caption{$\mathtt{NCI}$ improves the \textbf{overall performance} while \textbf{reducing latency} compared to strong baselines. 
        AUROC averaged across CIFAR-10 and ImageNet benchmarks in Table~\ref{tab:main}, with per image latency reported.
        }\label{tab:main_time}
    \end{subtable}
    \vspace{-7mm}
\end{table*}

\begin{table*}[t]
\vspace{-3mm}
\caption{
$\mathtt{NCI}$ reduces discrepencies and improves \textbf{overall performance} on ImageNet benchmarks across ViT B/16 and Swin v2 classifiers.
\textbf{Bold} marks best, \underline{underline} 2nd 
}\label{tab:vit}
\vspace{-5mm}
\begin{center}
    \centering
    \begin{subtable}{\linewidth}
        \begin{center}        \includegraphics[width=0.96\textwidth]{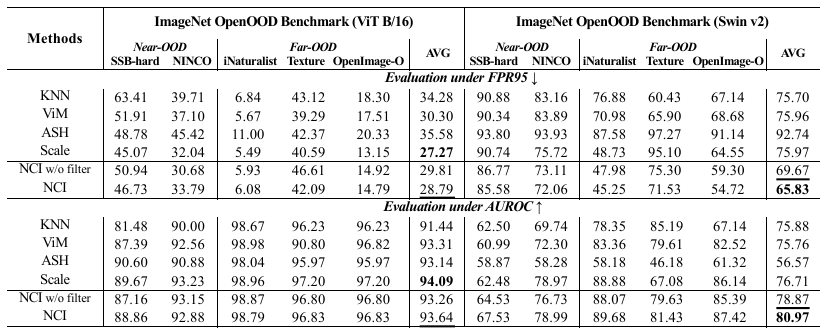} 
    \end{center}
        \centering
        \vspace{-3mm}
        \begin{minipage}{\linewidth}
        \centering
        \caption{
        $\mathtt{NCI}$ boosts Swin v2 while maintaining ViT effectiveness compared to baselines, even without filtering.
        }\label{tab:ViT}
        \end{minipage}
    \end{subtable}
    \begin{subtable}{\linewidth}
        \begin{center}        \includegraphics[width=0.65\textwidth]{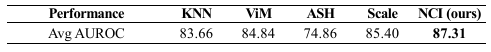} 
\end{center}
        \vspace{-3mm}
        \caption{$\mathtt{NCI}$ improves the \textbf{overall performance}. 
        AUROC averaged across two architectures in Table~\ref{tab:ViT}. 
        }\label{tab:vit_avg}
    \end{subtable}
\end{center}
\vspace{-6mm}
\end{table*}

In this section, we extensively evaluate $\mathtt{NCI}$ across classification tasks: CIFAR-10, CIFAR-100 (see App.~\ref{app:densenet}), ImageNet, as well as model architectures: ResNet, DenseNet (see App.~\ref{app:densenet}), ViT, Swin.
We compare $\mathtt{NCI}$ against \emph{thirteen} baseline methods. 
While $\mathtt{NCI}$ may not achieve the best performance on individual benchmarks, it mitigates the exisitng generalization discrepancies and achieves the best \textbf{overall} performance with \textbf{minimal} latency. 
Following the OpenOOD benchmark~\cite{zhang2023openood}, we evaluate on \emph{six} OOD sets for CIFAR-10 and CIFAR-100 classifiers and \emph{five} for ImageNet classifiers. 
Performance is evaluated using two widely recognized metrics: the False Positive Rate at 95\% True Positive Rate (FPR95) and the Area Under the Receiver Operating Characteristic Curve (AUROC).
Lower FPR95 and higher AUROC values indicate better performance.
We also report the per-image inference latency (in milliseconds) evaluated on a Tesla T4 GPU.
In our experiments, other than the ablation study in Section~\ref{sec:ablation}, we use the \(L1\)-norm as the filtering term and select the filtering strength $\alpha$ from $\{10^{-4}, 10^{-3}, 10^{-2}, 10^{-1}\}$ based on a validation set generated per pixel from Gaussian $N(0,1)$, following \cite{sun2021react,sun2022dice}.
Our method and all baselines are \emph{post-hoc} methods, while all models used are \emph{off-the-shelf} and do \emph{not} require complete Neural Collapse convergence.
For detailed setups and additional baselines, please see Appendix~\ref{app:imp-detail} and Appendix~\ref{app:additional_baselines}, respectively.

\subsection{Mitigating Discrepencies across ID Datasets}\label{sec:versatile}

We first assess $\mathtt{NCI}$ and baselines across CIFAR-10 and ImageNet classification tasks. 
The two tasks provide an ideal test bed for evaluating versatility, as they drastically differ in input resolution, number of classes, and classification accuracy. 
We use ResNets from OpenOOD \cite{zhang2023openood}: ResNet-18 for CIFAR-10 (95.06\% accuracy) and ResNet-50 for ImageNet (76.65\% accuracy).
Based on validation results, we set the filter strength $\alpha$ of the \( L1 \)-norm to $10^{-2}$ for CIFAR-10 experiments and $10^{-3}$ for ImageNet experiments.

\textbf{Datasets}
For CIFAR-10 experiments, We follow the OpenOOD split of ID test set and evaluate on the OpenOOD benchmarks, including CIFAR-100 \cite{krizhevsky2009learning}, Tiny ImageNet \cite{le2015tiny}, MNIST \cite{deng2012mnist}, SVHN \cite{netzer2011reading}, Texture~\citep{cimpoi2014describing}, and Places365~\citep{zhou2017places}. 
For ImageNet experiments, we follow the OpenOOD split of ID test set and evaluate on the OpenOOD benchmarks, including SSB-hard \cite{vaze2021open}, NINCO \cite{bitterwolf2023or}, iNaturalist~\citep{van2018inaturalist}, Texture~\citep{cimpoi2014describing}, and OpenImage-O~\cite{wang2022vim}. 

\textbf{Baselines}
In Table~\ref{tab:main}, we compare our method with \emph{thirteen} baselines.
Some baselines focus more on the CIFAR-10 Benchmark while others focus more focused on the Imagenet Benchmark.
Therefore, we categorize the baselines, besides the vanilla confidence-based $\mathtt{MSP}$ \citep{hendrycks2016baseline}, into two groups: the  ``CIFAR-10 Strong" baselines, including $\mathtt{ODIN}$ \citep{liang2018enhancing}, $\mathtt{Energy}$ \citep{liu2020energy}, $\mathtt{Mahalanobis}$ \citep{lee2018simple}, $\mathtt{KNN}$\citep{sun2022out}, $\mathtt{ViM}$ \citep{wang2022vim}, and $\mathtt{fDBD}$~\cite{liu2023fast}; 
the  ``ImageNet Strong" baselines, including $\mathtt{GradNorm}$ \citep{huang2021importance}, 
$\mathtt{NECO}$ \cite{ammar2023neco}, 
$\mathtt{React}$ \citep{sun2021react},  $\mathtt{Dice}$ \citep{sun2022dice},
$\mathtt{ASH}$ \cite{djurisic2022extremely}, 
$\mathtt{Scale}$ \cite{xu2023scaling}.
See Appendix~\ref{app:baseline} for details of the baselines.

\textbf{Performance} 
Table~\ref{tab:main} shows that $\mathtt{NCI}$ consistently ranks \emph{top-three} across benchmarks, whereas baselines exhibit greater variability. 
To assess overall performance, we averaged AUROC across benchmarks, which are of a similar range. 
Table~\ref{tab:main_time} highlights that NCI improves \emph{overall} performance compared to strong baselines on individual benchmarks.
Further, $\mathtt{NCI}$ is as efficient as $\mathtt{MSP}$, as shown in Table~\ref{tab:main_time}\footnote{Running time of $\mathtt{KNN}$ on ImageNet are copied from Table 4 in \cite{sun2022out}.}, which enhances efficiency compared to strong baselines.
This aligns with the analysis in Section~\ref{sec:methodology} and Appendix~\ref{app:baseline}.
We highlight the following pairs of comparison:

\begin{itemize}[left=0mm]
    \item \textbf{$\mathtt{NCI}$ v.s. $\mathtt{NCI}$ w/o filter:} 
    On the CIFAR-10 classifier, strong ID clustering allows our method to rank top-3 without filtering.
    Conversely, on the ImageNet ResNet-50, weaker ID clustering (see Appendix~\ref{app:nc_prevelance}) makes norm-based filtering crucial for reducing generalization discrepancy.
    Complete Neural Collapse occurs on neither model while $\mathtt{NCI}$ remains effective.
    
    \item \textbf{$\mathtt{NCI}$ v.s. $\mathtt{KNN}$}: Compared to $\mathtt{KNN}$, $\mathtt{NCI}$ significantly reduces the latency (Table~\ref{tab:main_time}). 
    Notably, without filtering, our hyperparameter-free score outperforms KNN with tuned parameters on most benchmarks (Table~\ref{tab:main}, Table~\ref{tab:ViT} \& Table~\ref{tab:dense_full}), highlighting the benefit of using class-specific information.  
    \item \textbf{$\mathtt{NCI}$ v.s. $\mathtt{ASH}$ / $\mathtt{Scale}$}: 
    Compared to both, $\mathtt{NCI}$ delivers competitive performance on ImageNet and \emph{significantly} improves CIFAR-10, enhancing \emph{overall} performance ( Table~\ref{tab:main_time}). 
    Also, $\mathtt{ASH}$ and $\mathtt{Scale}$ introduce in a small delay on the ImageNet benchmark due to activation sorting, with larger activation dimensions likely widening the latency gap on larger models.
    \vspace{0.5mm}
    \item \textbf{$\mathtt{NCI}$ v.s. $\mathtt{NECO}$:} 
    $\mathtt{NECO}$ \citep{ammar2023neco} is motivated by Neural Collapse.
    Like $\mathtt{NCI}$ with filtering, $\mathtt{NECO}$ uses max-logit and incorporates distance to the origin. 
    However, $\mathtt{NECO}$ exclusively analyzes features, requiring expensive matrix multiplication and leading to higher inference latency (Table~\ref{tab:main_time}).
    Conversely, $\mathtt{NCI}$ explores the interplay between features \emph{and} the classification head, integrating class-specific information to improve both efficiency and effectiveness.
\end{itemize}

\subsection{Mitigating Discrepancies across Architectures}\label{sec:versatile_transformer}

Next, we study two transformer-based models: ViT B/16 \cite{dosovitskiy2020image} and Swin-v2 \cite{liu2022swin}, both finetuned on ImageNet, achieving an accuracy of  81.14\% and 82.94\% respectively. 
We follow the setup of the OpenOOD ImageNet Benchmark in Section~\ref{sec:versatile}. 
Based on validation results, we set the filter strength $\alpha$ of the \(L1\) norm to $10^{-3}$ for both classifiers.
In Table~\ref{tab:vit}, we observe strong baselines suffer on Swin v2, echoing the observations in \cite{ammar2023neco}. 
Conversely, our $\mathtt{NCI}$, even without filtering, improves baseline performance on Swin v2. 
Filtering further enhances the performance, leading to improved \emph{overall} performance (Table~\ref{tab:vit_avg}).

\begin{table*}[t]
\vspace{-3mm}
\caption{\texttt{NCI} improves the \textbf{overall performance}, averaged across Table~\ref{tab:main}, Table~\ref{tab:ViT} \& Table~\ref{tab:dense_full}. 
}
\vspace{-5mm}
\label{tab:aggregated}
\begin{center}
\includegraphics[width=0.7\textwidth]{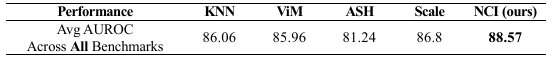} 
\end{center}
\vspace{-3mm}
\end{table*}

We further aggregate in Table~\ref{tab:vit} with experiments on ResNet (Table~\ref{tab:main_full}) and DenseNet (Table~\ref{tab:dense_full}) and report the average AUROC in Table~\ref{tab:aggregated}. 
$\mathtt{NCI}$ significantly boosts the overall performance.

\subsection{Ablation on the Filtering Effect}\label{sec:ablation}

In Table~\ref{tab:Lp_norm}, we assess different orders of $p$-norm as the filtering term, compared to the \( L1\) norm used so far. 
To ensure a fair comparison, we report 
the best performance from the filter strengths $\{10^{-4}, 10^{-3}, 10^{-2}, 10^{-1}\}$.
The rest of the setup follows the ImageNet benchmarks in Section~\ref{sec:versatile}. 
As shown in Table~~\ref{tab:Lp_norm}, filtering with \( L1\) norm achieves the best performance across OOD datasets, aligning with prior observations \cite{huang2021importance, park2023understanding}.
Meanwhile, we observe that in rare scenarios, e.g., a ResNet-18 on CIFAR-10, the $\mathtt{L1}$ norm cannot effectively characterize OOD's proximity to the origin, leading to no extra performance gain compared to simply thresholding on $\mathtt{pScore}$. 
In these cases, our algorithm benefits from its ability to automatically select a low filter strength based on validation results, effectively disregarding the filtering term.

\begin{table}[h]
\vspace{-2mm}
\caption{Ablation on filtering norm on ImageNet OpenOOD Benchmark with ResNet-50 backbone. 
AUROC score is reported (higher is better). 
\textbf{Bold} denotes the best.
Filtering with $\mathtt{L1}$ norm outperforms alternative choice of norms across OOD datasets.
}\label{tab:Lp_norm}
\vspace{-6mm}
\begin{center}
\includegraphics[width=0.49\textwidth]{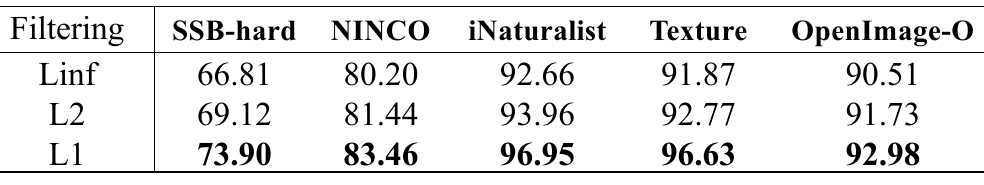}
\end{center}
\end{table}

We also test the sensitivity of $\mathtt{NCI}$ to filtering strength $\alpha$ in Table~\ref{tab:sensitivity}. 
As shown on the ImageNet ResNet50 benchmark, performance remains stable for $\alpha$ values within the same scale.
Given this insensitivity, we select hyperparameters from four scales $\{10^{-4}, 10^{-3}, 10^{-2}, 10^{-1}\}$ without extensive finetuning in this work.  

\begin{table}[h]
\vspace{-3mm}
\caption{
Sensitivity of $\mathtt{NCI}$ to filtering strength. %
Average AUROC on ImageNet ResNet-50 Benchmark reported. 
Performance remains stable within the same scale. 
}\label{tab:sensitivity}
\vspace{-6mm}
\begin{center}
\includegraphics[width=0.49\textwidth]{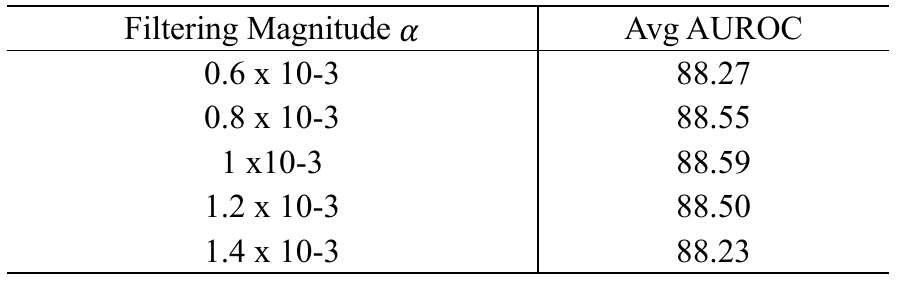}
\end{center}
\vspace{-4mm}
\end{table}

We further apply $\mathtt{L1}$-norm based filtering to $\mathtt{KNN}$ to see if this perspective can mitigate the discrepancy of clustering-based methods in general. 
In Table~\ref{tab:knn_l1} \footnote{
Note that we report our run of KNN here to ensure a fair evaluation of the filtering effect. 
Our results are very similar to the OpenOOD results reported in Table~\ref{tab:main}, with only marginal differences.
}, we report the the best performance of $\mathtt{KNN}$ from filter strengths $\{10^{-4}, 10^{-3}, 10^{-2}, 10^{-1}\}$.
We observe a significant performance gain from adding the filter, which further validates our understanding of ID clustering landscape from Neural Collapse.
Note that our method outperforms the standalone \(L1\) norm as well as $\mathtt{KNN}$, before and after filtering.

\begin{table}[h]
\caption{
Effectiveness of our filtering scheme on $\mathtt{KNN}$. Performance gain validates our understanding of ID clustering landscape. %
$\mathtt{NCI}$ outperforms $\mathtt{KNN}$ and standalone \(L1\) norm. 
AUROC reported (higher is better). 
\textbf{Bold} highlights the best result.
}\label{tab:knn_l1}
\vspace{-5mm}
\begin{center}
\includegraphics[width=0.49\textwidth]{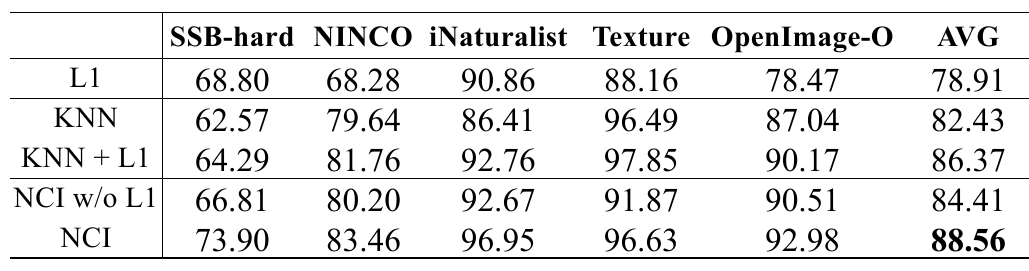}
\end{center}
\vspace{-6mm}
\end{table}

\section{Related Work}

\paragraph{OOD Detection}
Extensive research has focused on OOD detection algorithms.
One line of work is post-hoc and builds upon pre-trained models. 
For example, \cite{ hendrycks2019scaling, liang2018enhancing, liu2020energy, sun2021react, sun2022dice, liu2023fast, xu2024out} design OOD score over the output space of a classifier. 
Meanwhile, \cite{lee2018simple} and \cite{sun2022out} measure OOD-ness from the perspective of ID clustering in \emph{feature} space. 
Our work extends the observation that ID features tend to cluster from the perspective of Neural Collapse.
While existing work is more focused on certain classification tasks, our proposed OOD detector is tested to be highly versatile.

Others \citep{sharifi2024gradient, patil2024optimal, regmi2024t2fnorm, zhu2024croft} explore the regularization of OOD detection in training.
For example, \cite{ devries2018learning, hsu2020generalized} propose OOD-specific architecture whereas \cite{huang2021mos, wei2022mitigating} design OOD-specific training loss. 
\cite{tack2020csi} brings attention to representation learning for OOD detection and proposes an OOD-specific contrastive learning scheme.
Our work does not belong to this school of thought and is not restricted to specific training schemes or architecture. 
Notably, as shown in Appendix~\ref{app:t2f}, our NCI can also benefit from training-time algorithms.

\paragraph{Neural Collapse}
Neural Collapse was first observed in \cite{papyan2020prevalence}.
During Neural Collapse, the penultimate layer features collapse to class means, the class means and the classifier collapses to a simplex equiangular tight framework, and the classifier simplifies to adopt the nearest class-mean decision rule.
Further work has provided theoretical justification for the emergence of Neural Collapse \citep{han2021neural, mixon2020neural, zhou2022optimization, zhu2021geometric}.
Application-wise, \cite{zhu2021geometric} derives an efficient training algorithm inspired by Neural Collapse.
Our concurrent work \cite{ammar2023neco} also leverages Neural Collapse for OOD detection but overlooks class-specific information revealed by Neural Collapse, which is central to our approach.
\section{Conclusion}

This work leverages insights from Neural Collapse to propose a novel OOD detector. 
Specifically, we study the phenomenon that ID features tend to form clusters whereas OOD features reside far away.
Inspired by the trend of Neural Collapse prevalent on practical models, we hypothesize and validate that ID features tend to cluster near weight vectors. 
We also explain why ID features tend to reside further from the origin and complement our method from this perspective. 
Experiments show the effectiveness of our method on practical models without requiring the complete convergence of Neural Collapse. 
Further, our method improves the overall performance with minimal latency across diverse benchmarks. 
We hope our work can inspire future work to explore the interplay between features and weight vectors for OOD detection and other research problems such as calibration and adversarial robustness.
\section*{Acknowledgments}

We would like to express our sincere gratitude to Yubing Jian for conducting several experiments that made significant contributions to this work and for his valuable discussions.

{
    \small
    \bibliographystyle{ieeenat_fullname}
    \bibliography{main}
}

\clearpage
\setcounter{page}{1}
\maketitlesupplementary

\section{Implementation Details}\label{app:imp-detail}

\subsection{CIFAR-10}

\textbf{ResNet-18}
For visualization in Fig.~\ref{fig:demo-score}\textit{Left, Middle}, we use a CIFAR-10 classifier of ResNet-18 backbone trained with cross-entropy loss. 
The classifier is trained for 100 epochs, with the initial learning rate 0.1 decaying to 0.01, 0.001, and 0.0001 at epochs 50, 75, and 90 respectively.
For experiments in Table~\ref{tab:main}, we use the pre-trained model provided by the OpenOOD benchmark.
And we refer readers to \cite{zhang2023openood} for their training recipe.

\noindent \textbf{DenseNet-101}
For experiments on CIFAR-10 Benchmark presented in Table~\ref{tab:dense_full}, we evaluate a CIFAR-10 classifier of DenseNet-101 backbone. 
The classifier is trained following the setups in \cite{huang2017densely} with depth $L = 100$ and growth rate $k = 12$. 

\subsection{CIFAR-100}
\textbf{DenseNet-101}
For experiments on the CIFAR-100 Benchmark presented in Table~\ref{tab:dense_full}, we evaluate a CIFAR-100 classifier of the DenseNet-101 backbone. 
The classifier is trained following the setups in \cite{huang2017densely} with depth $L = 100$ and growth rate $k = 12$.

\subsection{ImageNet}

\textbf{ResNet-50}
For evaluation on ImageNet Benchmark in Table~\ref{tab:main}, we use the default ResNet-50 model trained with cross-entropy loss provided by Pytorch.
Training recipe can be found at
\url{https://pytorch.org/blog/how-to-train-state-of-the-art-models-using-torchvision-latest-primitives/}

\noindent \textbf{ViT B/16}
In Table~\ref{tab:vit}, we use the PyTorch implementation and pre-trained checkpoint of ViT B/16, available \url{https://github.com/lukemelas/PyTorch-Pretrained-ViT/tree/master}.

\noindent \textbf{Swin v2}
In Table~\ref{tab:vit}, we use the $\mathtt{timm}$ \cite{rw2019timm} implementation of Swin v2 as well as their pre-trained checkpoint 'swinv2\_base\_window8\_256'.

\section{Alternatives Proximity Metrics}\label{app:alter_metric}

In this section, we validate that under alternative similarity metrics, ID features also reside closer to weight vectors and empirically compare the metrics. 
In addition to our proposed $\mathtt{pScore}$,  we consider two standard similarity metrics, cosine similarity and Euclidean distance. 
For cosine similarity, we evaluate
\vspace{-2mm}
\begin{equation}
\mathtt{cosScore} = \frac{(\bm{h} - \bm{\mu}_G) \cdot \bm{w}_c}{\| \bm{h} - \bm{\mu}_G\|_2 \| \bm{w}_c \|_2}. 
\end{equation}

\vspace{-5mm}
As for Euclidean distance, we first estimate the scaling factor in Theorem~\ref{thm:1} by 
\(\displaystyle
\tilde{\lambda}_c = \frac{\| \bm{\mu}_{c} - \bm{\mu}_G\|_2}{\|\bm{w}_c\|_2}. 
\)
Based on the estimation, we measure the distance between the centered feature $\bm{h} - \bm{\mu}_G$ and the scaled weight vector corresponding to the predicted class $c$ as 
\vspace{-2mm}
\begin{equation}
\mathtt{distScore} = -\| ( \bm{h} - \bm{\mu}_G )- \tilde{\lambda}_c \bm{w}_c \|_2.
\vspace{-2mm}
\end{equation}
Same as $\mathtt{pScore}$, the larger $\mathtt{cosScore}$ or $\mathtt{distScore}$ is, the closer the feature is to the weight vector.

We evaluate in Table~\ref{tab:ablation_proximity} OOD detection performance using standalone $\mathtt{pScore}$, $\mathtt{cosScore}$, and $\mathtt{distScore}$ as scoring function respectively.
The experiments are evaluated with AUROC under the same ImageNet setup as in Section~\ref{sec:versatile}.
We observe in Table~\ref{tab:ablation_proximity}, that across OOD datasets, all three scores achieve an AUROC score $>50$, indicating that ID features reside closer to weight vectors compared to OOD under either metric. 

Furthermore, we observe that $\mathtt{pScore}$ outperforms both $\mathtt{cosScore}$ and $\mathtt{distScore}$. 
Comparing the performance of $\mathtt{pScore}$ and $\mathtt{cosScore}$, the superior performance of $\mathtt{pScore}$ implies that ID features corresponding to the classes with larger $\bm{w}_c$ are less compact. 
This is in line with the decision rule of the classifier that classes with larger $\bm{w}_c$ have larger decision regions. 
As for comparison against Euclidean distance based $\mathtt{distScore}$, $\mathtt{pScore}$ eliminates the need to estimate the scaling factor, which can be error-prone before convergence, potentially leading to performance degradation. 

\begin{table*}[t]
\caption{
Ablation on proximity scores.
AUROC score is reported (higher is better).
ID features are closer to weight vectors than OOD features (AUROC $>$ 50) under all metrics. 
Across OOD datasets, our proposed $\mathtt{pScore}$ can better separate ID an OOD features than $\mathtt{distScore}$ and $\mathtt{cosScore}$.}
\vspace{-6mm}
\label{tab:ablation_proximity}
\begin{center}
\includegraphics[width=0.75\textwidth]{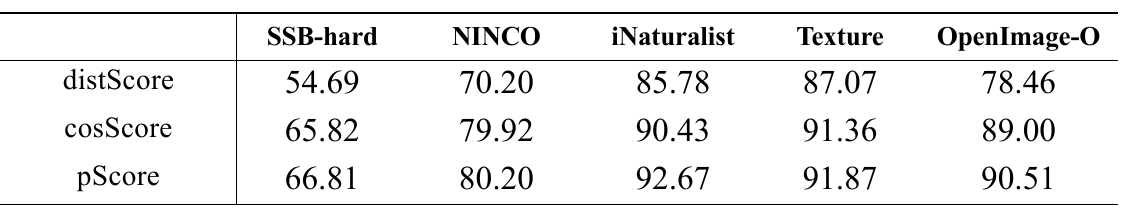}
\end{center}
\vspace{-6mm}
\end{table*}

\section{Baseline Methods}\label{app:baseline}

We provide an overview of our baseline methods in this session. 
We follow our notation in Section~\ref{sec:methodology}.
In the following, a lower detection score indicates OOD-ness.

\textbf{MSP} 
\cite{hendrycks2016baseline} proposes to detect OOD based on the maximum softmax probability. 
Given the penultimate feature $\bm{h}$ for a given test sample $\bm{x}$, the detection score of MSP can be represented as: 

\begin{equation}
\frac{\exp{(\bm{w}_c^T \bm{h}} + b_c)}{\sum_{c' \in \mathcal{C}} \exp{(\bm{w}_{c'}^T \bm{h}} + b_{c'})},
\end{equation}

where $c$ is the predicted class for $\bm{x}$.

\textbf{ODIN}
\cite{liang2018enhancing} proposes to amplify ID the OOD separation on top of MSP through temperature scaling and adversarial perturbation. 
Given a sample $\bm{x}$, ODIN constructs a noisy sample $\bm{x'}$ from $\bm{x}$. 
Denote the penultimate feature of the noisy sample $\bm{x'}$ as $\bm{h'}$, ODIN assigns OOD score following: 

\begin{equation}
\frac{\exp{((\bm{w}_c^T \bm{h'}} + b_{c})/T)}{\sum_{c' \in \mathcal{C}} \exp{((\bm{w}_c'^T \bm{h'}} + b_{c'})/T)},
\end{equation}

where $c$ is the predicted class for the perturbed sample and $T$ is the temperature. 
In our implementation, we set the noise magnitude as 0.0014 and the temperature as 1000. 

\textbf{Energy}
\cite{liu2020energy} designs an energy-based score function over the logit output.
Given a test sample $\bm{x}$ as well as its penultimate layer feature $\bm{h}$, the energy based detection score can be represented as: 

\begin{equation} 
- \log \sum_{c' \in \mathcal{C}} \exp{(\bm{w}_{c'}^T \bm{h}} + b_{c'}).
\end{equation}

\textbf{ReAct}
\cite{sun2021react} builds upon the energy score proposed in \cite{liu2020energy} and regularizes the score by truncating the penultimate layer estimation. 
We set the truncation threshold at $90$ percentile in our experiments.  

\textbf{Dice}
\cite{sun2022dice} builds upon the energy score proposed in \cite{liu2020energy}. 
Leveraging the observation that units and weights are used sparsely in ID inference, \cite{sun2022dice} proposes to select and compute the energy score over a selected subset of weights based on their importance.  
We set a threshold at $90$ percentile for CIFAR experiments and $70$ percentile for ImageNet experiments following \cite{sun2022dice}. 

\textbf{ASH} \cite{djurisic2022extremely} builds upon the energy score proposed in \cite{liu2020energy}. 
Prior to the Energy score, ASH sorts each feature to find the top-k elements, scales up the top-k elements, and sets the rest to zero. 
We note that in addition to the cost of Energy, ASH introduces a sorting cost of $O(P \log k)$, where $P$ is the penultimate layer dimension. 

\textbf{Scale} \cite{xu2023scaling} builds upon the energy score proposed in \cite{liu2020energy}. 
Prior to the Energy score, Scale sorts each feature to find the top-k elements and based on the statistics, scales all elements in the feature. 
We note that in addition to the cost of Energy, Scale also introduces a sorting cost of $O(P \log k)$, where $P$ is the penultimate layer dimension.

\textbf{Mahalanobis}
On the feature space, \cite{lee2018simple} models the ID feature distribution as multivariate Gaussian and designs a Mahalanobis distance-based score: 

\begin{equation}
\max_c - (\bm{e_x} - \hat{\bm{\mu}_c})^T \hat{\Sigma}^{-1} (\bm{e_x} - \hat{\bm{\mu}_c}),
\end{equation}
where $\bm{e_x}$ is the feature embedding of $\bm{x}$ in a specific layer, $\hat{\mu_c}$ is the feature mean for class $c$ estimated on the training set, and $\hat{\Sigma}$ is the covariance matrix estimated over all classes on the training set.

On top of the basic score, \cite{lee2018simple} also proposes two techniques to enhance the OOD detection performance.
The first is to inject noise into samples.
The second is to learn a logistic regressor to combine scores across layers. 
We tune the noise magnitude and learn the logistic regressor on an adversarial constructed OOD dataset.
The selected noise magnitude is 0.005 in both our ResNet and DenseNet experiments. 

\textbf{KNN} \cite{chen2020simple} proposes to detect OOD based on the k-th nearest neighbor distance between the normalized embedding of the test sample $\bm{z_x}/|\bm{z_x}|$ and the normalized training embeddings on the penultimate space. 
\cite{chen2020simple} also observes that contrastive learning helps in improving OOD detection effectiveness.

\textbf{GradNorm}
\cite{huang2021importance} extracts information from the gradient space to detect OOD samples. 
Specifically, \cite{huang2021importance} defines the OOD score function as the $\mathtt{L1}$ norm of the gradient of the weight matrix with respect to the KL divergence between the softmax prediction for $\bm{x}$ and the uniform distribution. 
\begin{equation}
\|\frac{\partial D_{KL} (\bm{u}\|softmax{f(\bm{x})})}{\partial\bm{W}}\|_1.
\end{equation}

\textbf{ViM}
\cite{wang2022vim} proposes to integrate class-specific information into feature space information by adding energy score to the feature norm in the residual space of the training feature matrix. 
The detection score is designed to be:
\begin{equation}
    \alpha \sqrt{\bm{h}^T \bm{R} \bm{R} \bm{h}},
\end{equation}
where $\bm{R} \in R^{P\times(P-D)}$ correspond to the residual after subtracting the $D-$dimensional principle space.
In the preparation stage, ViM requires evaluating the residual/null space from the training data, which is computationally expensive given the data volume. 
During inference, large matrix multiplication is required, resulting in a computational complexity of $O((P - D)^2)$. 

\textbf{NECO} is inspired by the ETF structure of Neural Collapse to utilize feature subspace for OOD detection.
The detection score is designed to be 
\begin{equation}
    \text{MaxLogit} \times \frac{\sqrt{\bm{h}^T \bm{P} \bm{P} \bm{h}}}{\sqrt{\bm{h}^T \bm{h}}},
\end{equation}
where $\bm{P} \in R^{P\times d}$ correspond to the $d-$dimensional principle space.
In the preparation stage, NECO requires evaluating the residual/null space from the training data, which is computationally expensive given the data volume. 
During inference, large matrix multiplication is required, resulting in a computational complexity of $O((d)^2 + P)$.

\textbf{fDBD} 
\cite{liu2023fast} proposes to detect OOD based on estimated feature distance to decision boundaries of class $c \in \mathcal{C}$ besides its predicted class $f(\bm{x})$: 
\begin{equation}\label{eq:closedForm}
\Tilde{D}_f(\bm{h}, c) = \frac{|(\bm{w}_{f(\bm{x})} - \bm{w}_c)^T \bm{h} + (b_{f(\bm{x})} - b_c)|}{\left\lVert \bm{w}_{f(\bm{x})} - \bm{w}_{c} \right\rVert_2},
\vspace{-3mm}
\end{equation}
The detection score is designed as
\begin{equation}
    \frac{1}{|\mathcal{C}| - 1} \sum_{\substack{c \in \mathcal{C}}, \ c \neq f(\bm{x}) } \frac{\Tilde{D}_f(\bm{h}, c)}{\| \bm{h} - \bm{\mu}_{train} \|_2}. 
\end{equation}
fDBD has time complexity $O(|\mathcal{C}| + P)$, where $|\mathcal{C}|$ is the number of training classes and $P$ is the penultimate layer dimension. 


\section{Performance Boosting with Training-Time Regularization}\label{app:t2f}

In this section, we investigate the compatibility of NCI with training-time regularization algorithms for OOD detection. 
On the CIFAR-10 benchmark, we evaluate NCI performance on models trained with a training-time regularization method, T2FNorm\cite{regmi2024t2fnorm}, as an example. 
We also compare this with NCI performance on models trained using standard cross-entropy loss, as presented in Table~\ref{tab:main_full}. 
In Table~\ref{tab:t2f}, we report the average AUROC across near OOD datasets (CIFAR-100 and TIN) and far OOD datasets (MNIST, SVHN, Texture, and Place365). 
We observe a performance boost in both cases, demonstrating the compatibility of NCI as a post-hoc method with training-time regularization algorithms. 
This also highlights the effectiveness of combining NCI with training-time regularization for improved OOD detection performance.

\section{Additional Baselines}\label{app:additional_baselines}

In the setup of Section~\ref{sec:versatile}, we further compare our NCI with two additional baselines, GEN \cite{liu2023gen} and SHE \cite{zhang2022out}.
On the CIFAR-10 benchmark, SHE, GEN, and NCI achieve average AUROCs of 84.06, 90.30, and 90.46, respectively.
On the ImageNet benchmark, SHE, GEN, and NCI achieve average AUROCs of 84.06, 86.20, and 88.56.
Our result further demonstrates NCI’s superior performance in mitigating generalization dependency across different classification tasks.

In the setup of Section~\ref{sec:versatile_transformer}, we apply the CLIP-based method MCM \cite{ming2022delving} (for vision-language models) to the vision models studied in that section. 
For MCM, we report the best average AUROC scores achieved through temperature sweeping across [0.01,0.1,1,10] for each model. 
On ViT, MCM achieves an average AUROC of 93.60, whereas NCI achieves 93.64. 
On Swin v2, MCM achieves an average AUROC of 77.73, whereas NCI achieves 80.94, demonstrating NCI's superior performance in this setup.

\begin{table}[h]
\caption{
Performance Boosting with Training-Time Regularization. 
On the CIFAR-10 benchmark, we report the average AUROC for Near and Far OOD. 
Experiments are conducted with ResNet-18 trained using Cross-Entropy (CE) and T2FNorm. 
}\label{tab:t2f}
\vspace{-5mm}
\begin{center}
\includegraphics[width=0.48\textwidth]{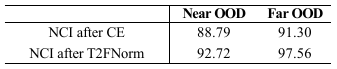}
\end{center}
\vspace{-3mm}
\end{table}

\section{Evaluation on DenseNet}\label{app:densenet}
In addition to evaluation on ResNet and transformer-based model in Section~\ref{sec:experiments}, we report the performance of our $\mathtt{NCI}$ along with the baselines under AUROC and FPR95 across OpenOOD benchmarks in Table~\ref{tab:dense_full}.

\begin{table*}[t]
\caption{\textbf{Our OOD detectors achieves high AUROC and low FPR95 across CIFAR-10 and CIFAR-100 OOD benchmark on DenseNet.}
$\uparrow$ indicates that larger values are better and vice versa.
\textbf{Bold} highlight the best results and \underline{underline} denotes the 2nd and 3rd best results.
We note that for DenseNet CIFAR-10 and CIFAR-100 classifiers, the discrepancy among existing methods is not as severe as in the examples presented in the main paper. Nevertheless, our $\mathtt{NCI}$ achieves state-of-the-art performance or improves upon existing methods, enhancing overall performance on average.
}
\vspace{-2mm}
\label{tab:dense_full}
\begin{center}
\includegraphics[width=1\textwidth]{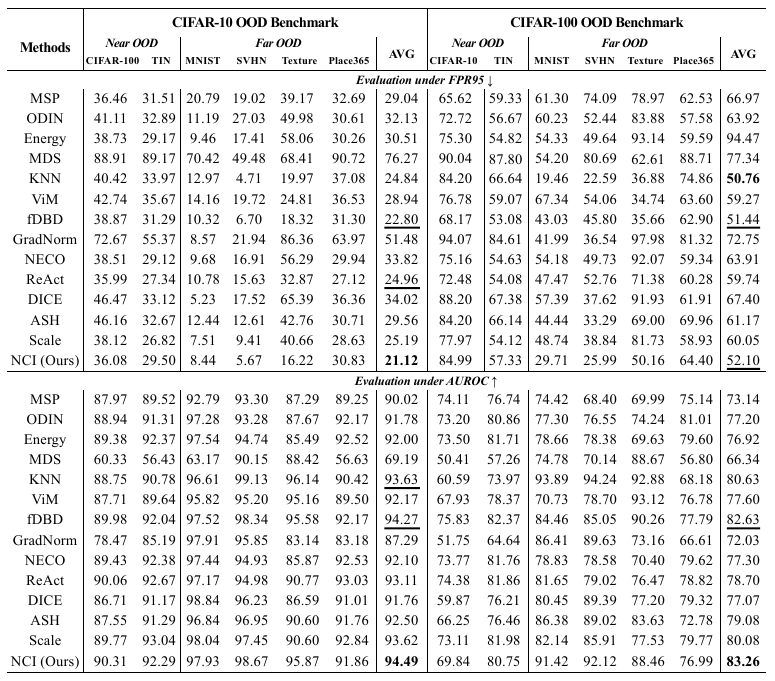} 
\end{center}
\vspace{-6mm}
\end{table*}

\section{The Prevalence of Neural Collapse across Canonical Classification Tasks }\label{app:nc_prevelance}

\begin{figure*}[h]
\begin{center}
\centerline{\includegraphics[width=\textwidth]{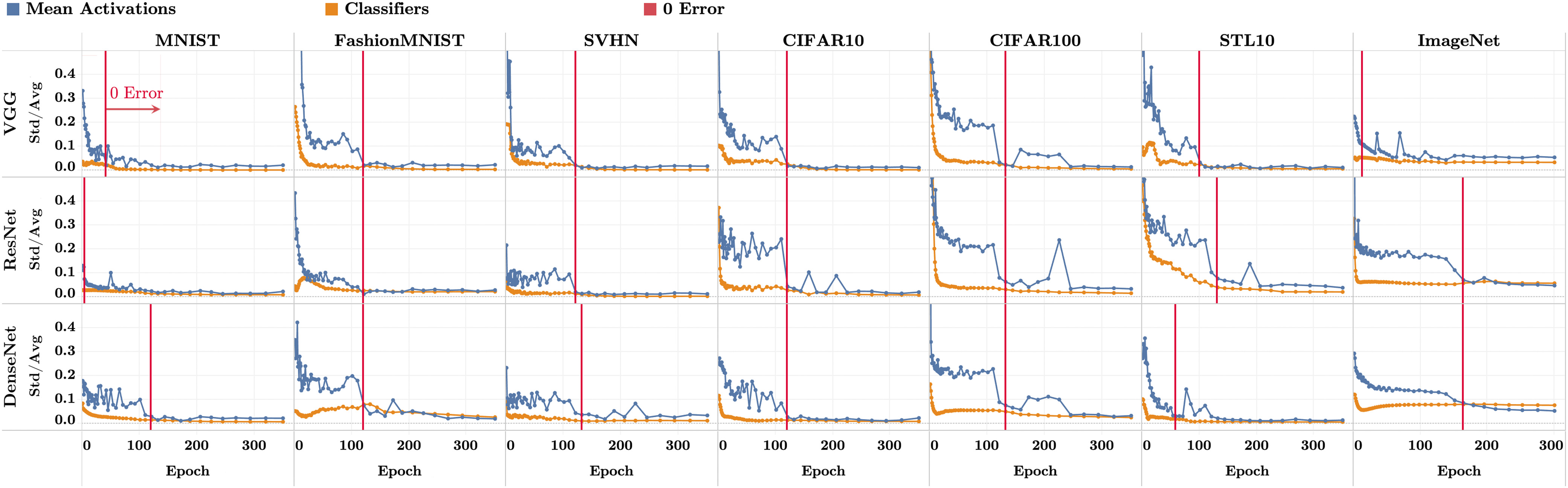}}
\end{center} 
\vspace{-8mm}
\caption{
(ref. Figure 2 in \cite{papyan2020prevalence})
\textbf{Train class means become equinorm.} 
In each array cell, the vertical axis shows the coefficient of variation of the centered class-mean norms as well as the network classifiers norms. In particular, the blue lines show $\text{Std}_c(\|\bm{\mu}_c - \bm{\mu}_G \|_2)/\text{Avg}(\|\bm{\mu} - \bm{\mu}_G\|_2)$ where $\{\bm{\mu}_c \}$ are the class means of the last-layer activations of the training data and $\bm{\mu}_G$ is the corresponding train global mean; the orange lines show$\text{Std}_c(\|\bm{w}_c\|_2)/\text{Avg}(\|\bm{w}_c\|_2)$ where $\{\bm{w}_c \}$ is the last-layer classifier of the $c$~th class. As training progresses, the coefficients of variation of both class means and classifiers decrease.
}\label{fig:nc_fig02}
\vspace{-5.5mm}
\end{figure*}

\begin{figure*}[h]
\begin{center}
\centerline{\includegraphics[width=\textwidth]{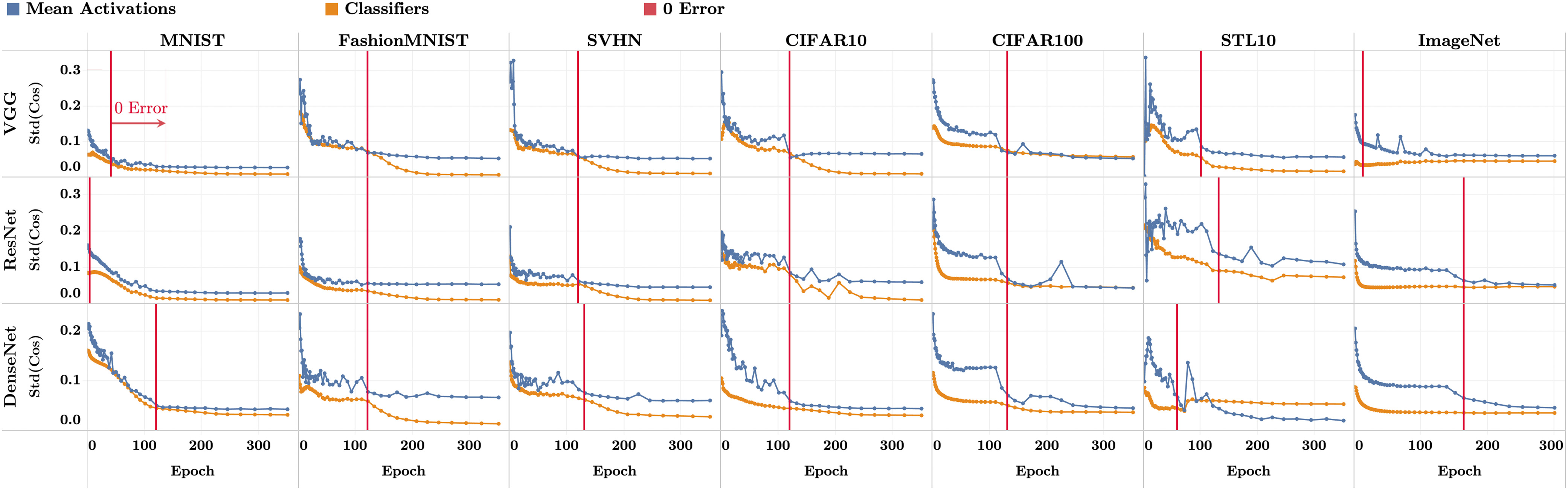}}
\end{center} 
\vspace{-8mm}
\caption{
(ref. Figure 3 in \cite{papyan2020prevalence})
\textbf{Classifiers and train class means approach equiangularity.} 
In each array cell, the vertical axis shows the SD of the cosines between pairs of centered class means and classifiers across all distinct pairs of classes $c$ and $c'$. Mathematically, denote
$\cos_\mu(c, c') = <\bm{\mu}_c - \bm{\mu}_G, \bm{\mu}_c' - \bm{\mu}_G> / \|\bm{\mu}_c - \bm{\mu}_G\|_2 \|\bm{\mu}_c' - \bm{\mu}_G\|_2 $
and 
$\cos_w(c, c') = <\bm{w}_c, \bm{w}_c'> / \|\bm{w}_c\|_2 \|\bm{w}_c'\|_2$,
where $\{\bm{w}_c\}_{c = 1}^C, \{\bm{\mu}_c\}_{c = 1}^C$, and $\bm{\mu}_G$ are as in Figure~\ref{fig:nc_fig02}.
We measure $\text{Std}_{c, c'}(\cos_\mu(c, c'))$ (orange) and $\text{Std}_{c, c'}(\cos_w(c, c'))$. As training progresses, the SDs of the cosines approach zero, indicating equiangularity.
}\label{fig:nc_fig03}
\vspace{-5.5mm}
\end{figure*}

\begin{figure*}[h]
\begin{center}
\centerline{\includegraphics[width=\textwidth]{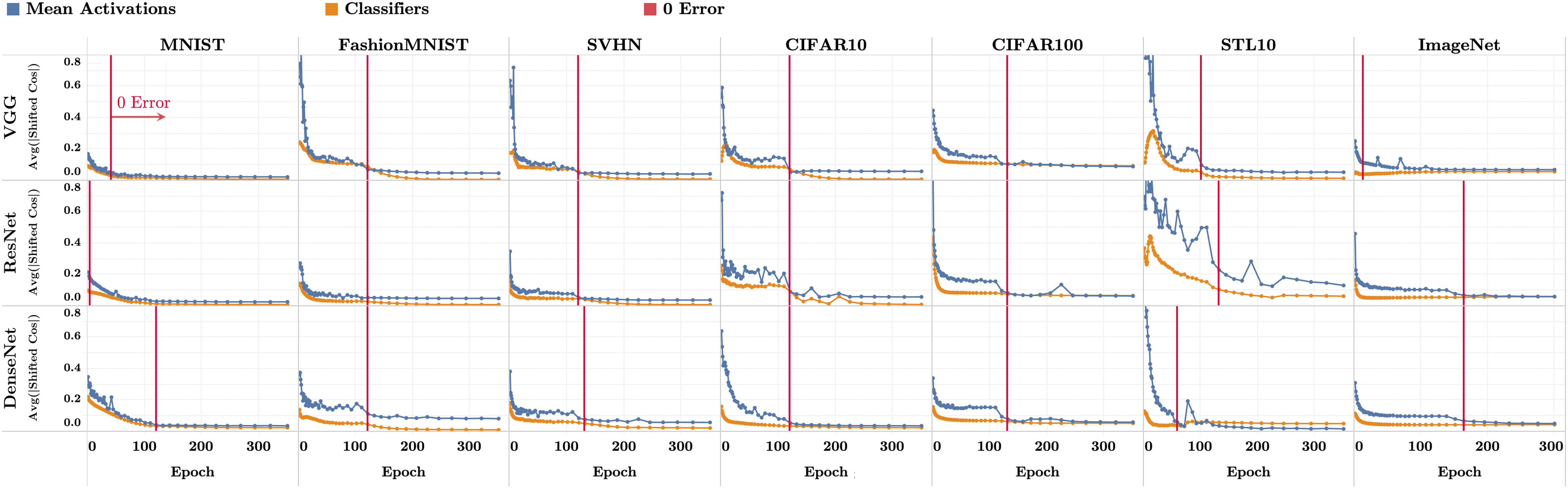}}
\end{center} 
\vspace{-8mm}
\caption{
(ref. Figure 4 in \cite{papyan2020prevalence})
\textbf{Classifiers and train class means approach maximal-angle equiangularity.} 
We plot in the vertical axis of each cell the quantities $\text{Avg}_{c, c'} |\cos_\mu(c, c') + 1/(C -1)|$ (blue) and $\text{Avg}_{c, c'} |\cos_w (c, c') + 1/(C -1)|$ (orange), where $\cos_\mu (c, c')$ and $\cos_w (c, c')$ are as in Figure~\ref{fig:nc_fig03}. 
As training progresses, the convergence of these values to zero implies that all cosines converge to 
$-1/(C-1)$. 
This corresponds to the maximum separation possible for globally centered, equiangular vectors.
}\label{fig:nc_fig04}
\vspace{-5.5mm}
\end{figure*}

\begin{figure*}[h]
\begin{center}
\centerline{\includegraphics[width=\textwidth]{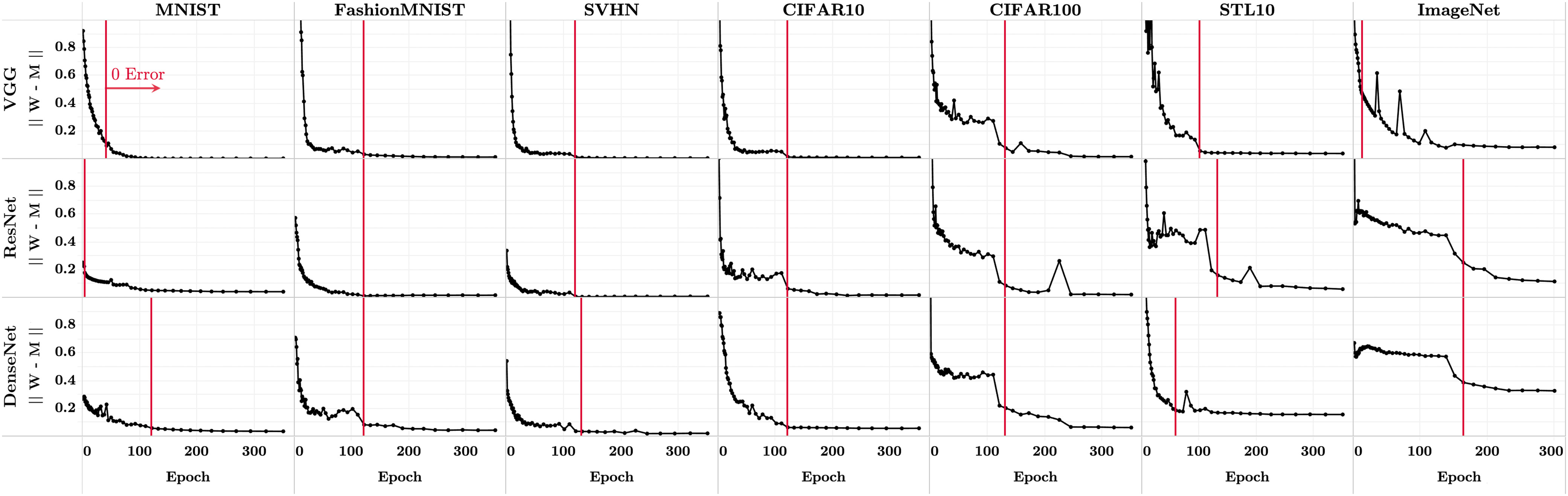}}
\end{center} 
\vspace{-8mm}
\caption{
(ref. Figure 5 in \cite{papyan2020prevalence})
\textbf{Classifier converges to train class means.} 
The formatting and technical details are as described in Section 3. In the vertical axis of each cell, we measure the distance between the classifiers and the centered class means, both rescaled to unit norm. Mathematically, denote $\Tilde{\bm{M}} = \Dot{\bm{M}} / \| \Dot{\bm{M}} \|_F$ where $\Dot{\bm{M}} = [\bm{\mu}_c - \bm{\mu}_G, c = 1, ...., C] \in \mathtt{R}^{P \times C}$ is the matrix whose columns consist of the centered train class means; denote $\Tilde{\bm{W}} = \bm{W}/\| \bm{W} \|_F$ where $\bm{W} \in \mathtt{R}^{C \times P}$ is the last-layer classifier of the network. We plot the quantity $\|\Tilde{\bm{W}}^T - \Tilde{\bm{M}} \|_F^2$ on the vertical axis. This value decreases as a function of training, indicating that the network classifier and the centered-means matrices become proportional to each other (self-duality).
}\label{fig:nc_fig05}
\vspace{-5.5mm}
\end{figure*}

\begin{figure*}[h]
\begin{center}
\centerline{\includegraphics[width=\textwidth]{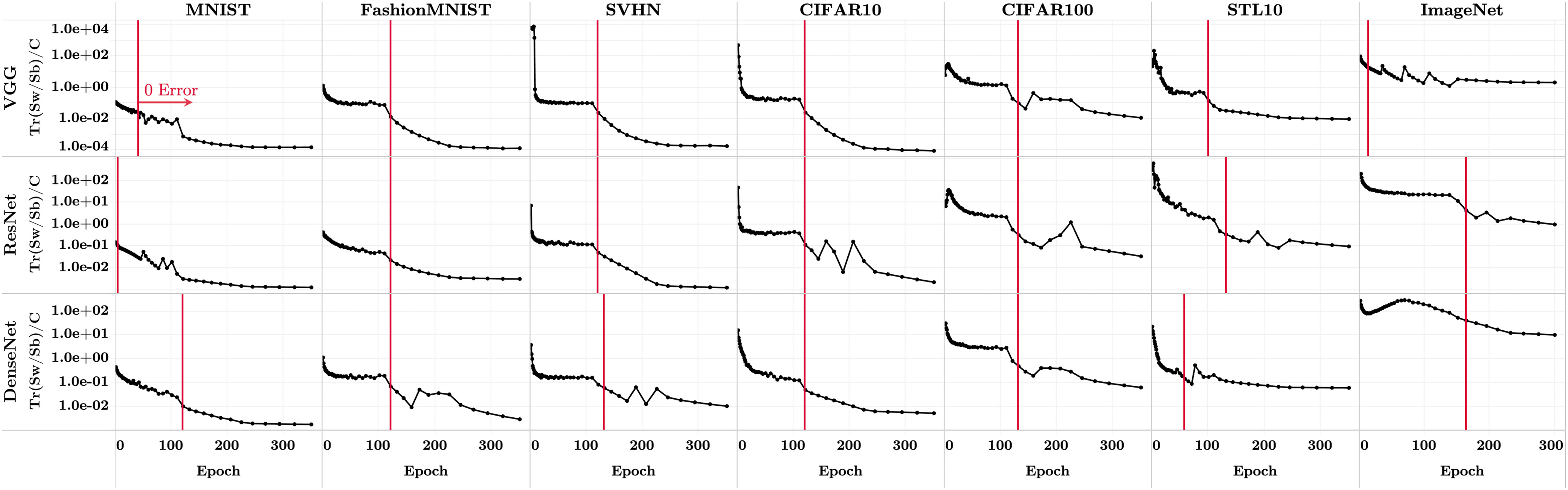}}
\end{center} 
\vspace{-8mm}
\caption{
(ref. Figure 6 in \cite{papyan2020prevalence})
\textbf{Training within-class variation collapses.} 
In each array cell, the vertical axis (log scaled) shows the magnitude of the between-class covariance compared with the within-class covariance of the train activations. Mathematically, this is represented by $\text{Tr}(\Sigma_W\Sigma_B^+/C)$ where $\text{Tr}(\cdot)$ s the trace operator, $\Sigma_W$ is the within-class covariance of the last-layer activations of the training data, $\Sigma_B$ is the corresponding between-class covariance, $C$ is the total number of classes, and $[\cdot]^+$ is Moore–Penrose pseudoinverse. This value decreases as a function of training—indicating collapse of within-class variation.
}\label{fig:nc_fig06}
\vspace{-5.5mm}
\end{figure*}

The phenomenon of Neural Collapse, as established in the seminal work by Papyan et al. \cite{papyan2020prevalence} and corroborated by subsequent studies \cite{han2021neural, mixon2020neural, zhou2022optimization, zhu2021geometric}, widely exists across canonical classification datasets and model architectures. 
The prevalent occurrence of Neural Collapse forms a robust foundation for the design of our versatile OOD detectors. 
To this end, we review the empirical evidence of Neural Collapse across different datasets and model architectures in Figure~\ref{fig:nc_fig02}, Figure~\ref{fig:nc_fig03}, Figure~\ref{fig:nc_fig04},  
Figure~\ref{fig:nc_fig05},
and Figure~\ref{fig:nc_fig06}.
Comparing CIFAR-10 and ImageNet behaviors with ResNet backbone in Figure~\ref{fig:nc_fig06}, we note that the clustering of CIFAR-10 is more prominent than Imagenet, as indicated by a higher ratio of between-class variance to within-class covariance. 
Note that the figures and captions are sourced from  \cite{papyan2020prevalence}.
The definition and notation follow Section~\ref{sec:methodology}. 

\end{document}